**Short Paper**

# Predicting Pollution Level Using Random Forest: A Case Study of Marilao River in Bulacan Province, Philippines

Jayson M. Victoriano
AMA University, Bulacan State University, Philippines
jayson.victoriano@bulsu.edu.ph
(corresponding author)

Manuel Luis C. Delos Santos
Asian Institute of Computer Studies, Philippines
saintawee@gmail.com

Albert A. Vinluan
New Era University, Philippines
aavinluan@neu.edu.ph

Jennifer T. Carpio
University of the East, Philippines
jhennieffer.carpio@gmail.com

## Abstract

*Purpose* – This study aims to predict the pollution level that threatens the Marilao River, located in the province of Bulacan, Philippines. The inhabitants of this area are now being exposed to pollution. Contamination of this waterway comes from both formal and informal industries, such as a used lead-acid battery, open dumpsites metal refining, and other toxic metals. Using various water quality parameters like Dissolved Oxygen (DO),

Potential of Hydrogen (pH), Biochemical Oxygen Demand (BOD) and Total Suspended Solids (TSS) were the basis for predicting the pollution level.

*Method* – This study used the Data Mining technique based on the sample data collected from January of 2013 to November of 2017. These were used as a training data and test results to predict the river condition with its corresponding pollution level classification indicated with the used of colors such as "Green" for "Normal", "Yellow" for "Average", "Orange" for "Polluted" and "Red" for "Highly Polluted". The model got an accuracy of 91.75% with a Kappa value of 0.8115, interpreted as "Strong" in terms of the level of agreement.

*Results* – The predicted model using the Random Forest have scored 91.75% in terms of correctly classified instances and were able to generate 0.8115 Kappa values which indicate that the model used to produce a strong level of agreement.

*Conclusion* – From 2013 to 2017 based on the data sampling provided by the Environmental Management Bureau (EMB), an attached agency of the Department of Environment and Natural Resources (DENR) in the Philippines mandated to protect and restore the environment, shows that the river is highly polluted. Several issues like, underestimation of the water parameter results have been identified, issues which can be addressed by incorporating more observations to the training process and by validating the resulting model on the different training set. The discretion on decisions about the prediction of the model is attributed to DENR-EMB unit as they have more hands-on experience with regards to monitoring, restoring, protecting the environment.

## INTRODUCTION

Marilao River is included in the "World's Worst Polluted Places" as reported by PureEarth formerly known as Blacksmith Institute (Blacksmith Institute, 2007). Figure 1 shows the observed condition of Marilao River as the result of river quality monitoring since 2005. Aside from organic pollution, there were abundances of heavy metal that may pose significant health risks to surrounding communities that depend on the river system. According to (Malenab et al. 2016), heavy metal pollution came from used jewelry smelting, tanneries, used lead-acid battery recycling and other industries dealing with heavy metals commonly in the upstream area of the river system.

These pollutants, particularly the significant metals, create a major health risk to close communities that surround the stream water for fish ponds, bathing, and swimming that causes some health issues. Additionally, 31% of all diseases within the country are attributed to contaminated waters. An average of 55 Filipinos per day suffers from diseases attributed to poor sanitation and poor water quality (Malenab et al., 2016). Figure 2 shows the map of Marilao River covering around 20 km from upstream Caloocan down to Meycauayan and Obando. In the 2015 census, the said river has a population of 221, 965 people which will be the affected person. With the use of data mining methods and random forest classifier, it can be determined in advance the water pollution level of Marilao River and its pattern to offer prediction ahead of time. The data (water parameter result) Dissolve Oxygen(DO), Potential Hydrogen (PH), Biochemical Oxygen Demand(BOD) and Total Suspended Solids (TSS) that are being collected from January 2013 to November 2017 will be used as a training set and test set to decide the accuracy of the predictive model and help the local government unit of Marilao and its municipal communities in providing awareness and consciousness about the pollution level of the river.

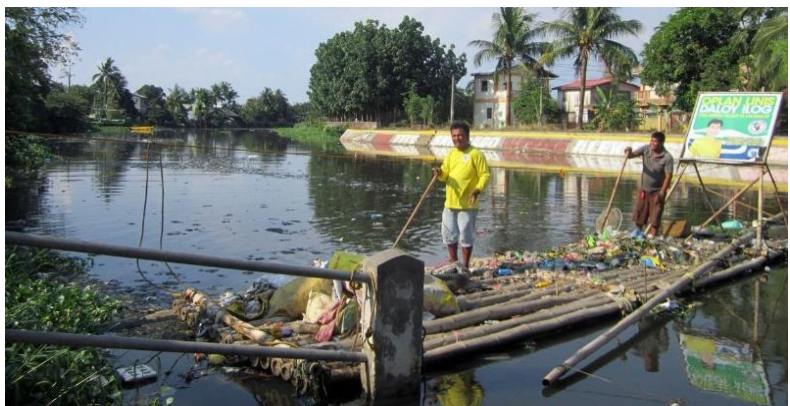

*Figure 1.* Water Condition around Marilao River

There are two main objectives to be solved in this study. The first is at the best prognostic features influencing the prediction, and the second—building the structure of the predicting system which provides the most accurate forecast.

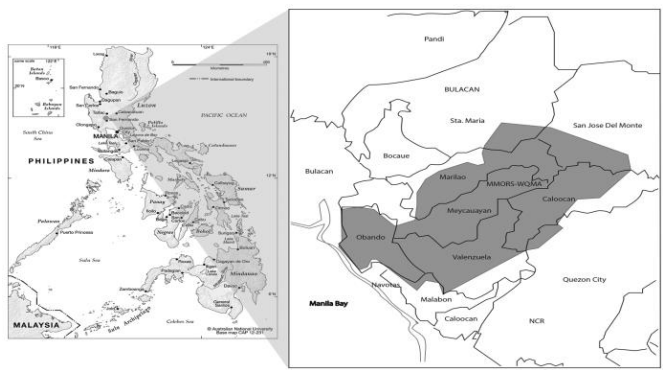

*Figure 2.* Map of the Marilao-Meycauayan-Obando River System, Water Quality Management Area - MMORS WQMA (Amparo et al. 2017)

## LITERATURE REVIEW

This section highlights several studies for water quality prediction models. These studies on water and river system prediction were done based on the study of Cagayan River under the Department of Science and Technology (DOST) give an ahead of time forecasting system for water level and flood hazard specifically on monitoring stations situated along the Cagayan River Basin the study includes using a predictive model using Random Forest. In the river studied, flood forecasting was based mainly on linear regression equations for water level observations, which do not consider downstream rainfall effects. Forecasting lead time, forecasting accuracy, and issuing and disseminating warnings are the three critical components of flood forecasting and early warning. Thus, predicting the onset of flooding accurately is very important to reliable flood forecasting. the resulting accuracy of the model produces an average above 0.90 and indicate high prediction accuracy (Garcia, Retamar, & Javier 2016).

Environmental problems, including the degradation and depletion of natural resources, biodiversity loss, and climate change, among others, represent some of the most critical challenges of our world today. To effectively address environmental problems, understanding how these system components affect one another is needed. Data Science (DS) is an emergent research field that helps better understand the complex mechanisms behind environmental phenomena (Gibert et al., 2018).

A study also shows that surface water bodies in general, and the rivers, in particular, are among the most vulnerable aquatic systems to contamination due to their easy accessibility for the disposal of wastes (Yuan 2016). The author described that the river water pollution foremost affects its chemical quality and then systematically deteriorates the community disrupting the food web, river pollution has various dimensions that must the ability (Chawla, 2015).

Prediction of water quality is a way to study the future status of water quality by using some prior knowledge and data. Modeling involves water quality; decision makers can understand the tendency of water quality in current and future periods (Jian-jun, Chuan-biao, & Ming-hua, 2010).

## METHODOLOGY

Random forest theory develops rapidly in the front fields of the international researches in recent years, its development has a major impact on the computer, artificial intelligence, and machine learning. Data mining is the process of analyzing large information repositories and of discovering implicit (Han, Kamber, & Pei, 2012), but potentially useful information there as some process underpinning to come up with the knowledge discovery those are selecting the target data followed by processing of data for the specific need which is transforming data, this process involved eliminating noise and outlier results to produce trends and pattern and produce knowledge discovery shown in Figure 3.

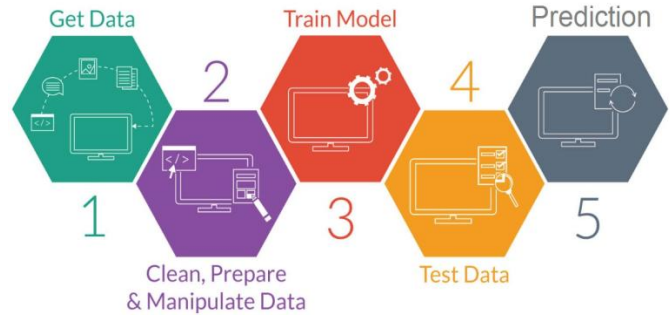

*Figure 3.* Predictive Modelling Phases

Random Forest (RF) is an ensemble learning algorithm that creates many shallow uncorrelated decision tree predictors based on data sampled from the training set. The idea is that by assembling a set of weak learners, a strong learner is formed (Au, 2017).

---

Algorithm: Random Forest

Input: Let X be the training data consisting of L variable feature vectors. Let B be the amount of trees in a Random Forest.

Random Forest Training
1. For *i* =1, . . . . , B, iterate until convergence:
 (a) Draw a bootstrap sample S of size N from X.
 (b) Grow a tree Tb from the bootstrapped data with the
   following conditions:
 i. Given the L input variable, a number *l* << L is specified
   such that for each node. *l* variables are selected
   randomly from X and the best split from l is used to
   split the node.
 ii. Grow the tree without pruning
2. Output the ensemble of trees

Random Forest Prediction
Let x be a feature vector of a test data, the
prediction is given by: $\{T_b\}\frac{B}{1}$

3. $\hat{f}\frac{B}{rf}(x) = \frac{1}{B}\sum_{i=1}^{B} T_b(x)$

---

*Figure 4.* Random Forest Algorithm during the training process and testing data

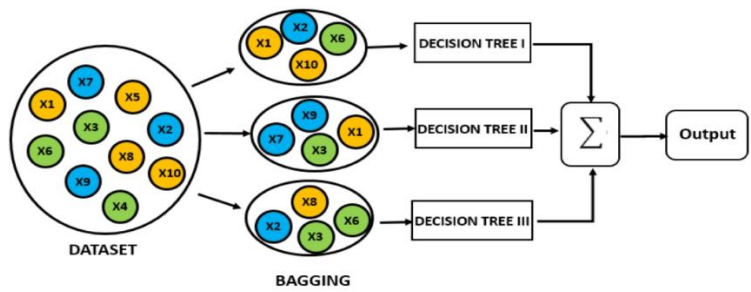

*Figure 5.* Overview of the Random Forest Algorithm

The Random Forest algorithm process as shown in Figure 4 and Figure 5, Random Forest works by building decision trees from a bootstrapped sample taken from a training set. This process is repeated B several times where B is the desired number of trees generated for the forest (Garcia, Retamar, & Javier 2016). During the construction of a tree, a node is split based on the best among the random subset of the features. While the introduction of randomness results into an increasing bias of the prediction, by averaging the results of each generated tree, the resulting prediction of the Random Forest is ensured to have low bias and variance (Breiman, 2001).

## Data Preparation

The collected dataset is based on the Department of Environment and Natural Resources (DENR) Environment Management Board (EMB) Region 3 starting from January of 2013 to November 2017 this includes the different water parameter and standard show in Table 1. The authors focus on DO, pH, BOD, and TSS which measures an approximate amount of biodegradable organic matter present in water is generally used as the criterion to measure in determining the water quality of the river. Based on Water Quality Guideline of DENR Administrative Order, (Paje & Cuna, 2016) Marilao River is classified as Type C body of water.

Table 1.  Water Quality Parameters

| Parameter | Standard |
|-----------|----------|
| BOD | 5 (mg/l) |
| pH | 6.5-8.5 (mg/l) |
| TSS | 65 (mg/l) |
| DO | 5 (mg/l) |

## Feature Engineering

The cleaned data contains the following columns:

a.) BOD (mg/l)
b.) PH ((mg/l)
c.) TSS(mg/l)
d.) DO(mg/l)
e.) Pollution Level

## Training of the Prediction Model

This section describes what tools the researcher used in training and the supplied training data. This paper applies the Waikato Environment for Knowledge Analysis (WEKA) data analytic software and Orange Visual programming for visualization. Random Forest classifier was also used during the training process to provide a prediction with 10-fold cross-validation to avoid overfitting and to get a more accurate result. Collected data started from January 2013 up to July 2016 with a total of 473 instances.

## Validation

This section describes the different metrics used by the researcher in evaluating the classifier model performance; its effectiveness and the quality of its prediction. Several tests of data with known water quality parameter values were used to test the accuracy of the generated sample by distinguishing the reliability of the data and their validity in accordance to the comparison of an observed accuracy with an expected accuracy rate that is likely to meet based on the Confusion Matrix (Mallari et al., 2018). The classifier can also be evaluated in terms of precision, recall and F-measure and the assessment of interrater reliability (Vinluan et al., 2018). Cohen's Kappa is used for validation purposes which measure the interrater reliability between the observed accuracy with an expected accuracy shown in Table 2.

Precision is the ratio of relevant instances in the retrieved instances that are referred to as a positive value. Precision was calculated as shown in equation 1 where $tp$ is truly positive and $fp$ is a false-positive.

$$Precision = tp/(tp+fp) \qquad Equation\ 1$$

Recall it is defined as the true positive rate to calculate recall equation 2 must be used, where $tp$ is true positive and $fn$ is a false negative.

$$Recall = tp/(tp+fn) \qquad Equation\ 2$$

F-measure is the weighted average of Precision and Recall and is calculated as shown in equation 3.

$$F\ Score = 2*(Recall * Precision) / (Recall + Precision) \qquad Equation\ 3$$

Interrater Reliability is to measure interrater reliability between two raters, Cohen's Kappa statistic is used which is shown in equation 4, where $P_o$ is the relative observed agreement among raters, $P_e$ is the hypothetical probability of chance agreement and K is the Kappa value.

$$K = (P_o - P_e) / (1 - P_e) \qquad Equation\ 4$$

Table 2. Kappa Value and level of agreement (Landis and Koch 1977)

| Value of Kappa | Level of Agreement |
|---|---|
| 0.00-0.20 | None |
| 0.21-0.39 | Minimal |
| 0.40-0.59 | Weak |
| 0.60-0.79 | Moderate |
| 0.80-0.90 | Strong |
| Above 0.90 | Almost Perfect |

## RESULTS AND DISCUSSION

This section discusses the result and finding of the researcher based on the data mining process. The selected test data from August 2016 to November 2017 onwards were consists of 73 instances used as testing data. The correctness of the classifier is measured using the WEKA tool of which the results are shown below in Table 4. The detailed accuracy by class is presented in Table 5 and the Confusion Matrix shown in Table 6. Pollution level classification is shown in Table 3 of Marilao River together its corresponding equivalent value.

Table 3. Marilao River Pollution Level Classification

| River Condition | Pollution Level Classification |
|---|---|
| Normal | Green |
| Average | Yellow |
| Polluted | Orange |
| Highly Polluted | Red |

The normal condition which is colored "Green" means, that the rivers system is habitable to all living things and organisms. The average condition that is colored "Yellow" shows some signs of inappropriate unwanted toxins found in the river, polluted condition that is colored "Orange" indicates hazardous toxins to all its surrounded communities and the highly polluted river that is colored "Red" is considered dead and not suitable for all living organisms because it contains high level of toxic and unwanted minerals.

The use of Scatter Plotter is to visualize how the training set and its predictive model evidently shown in Figures 6-9, how closely fit the observed value and the predictive value in the study and the model is able to follow the general trend in pollution level.

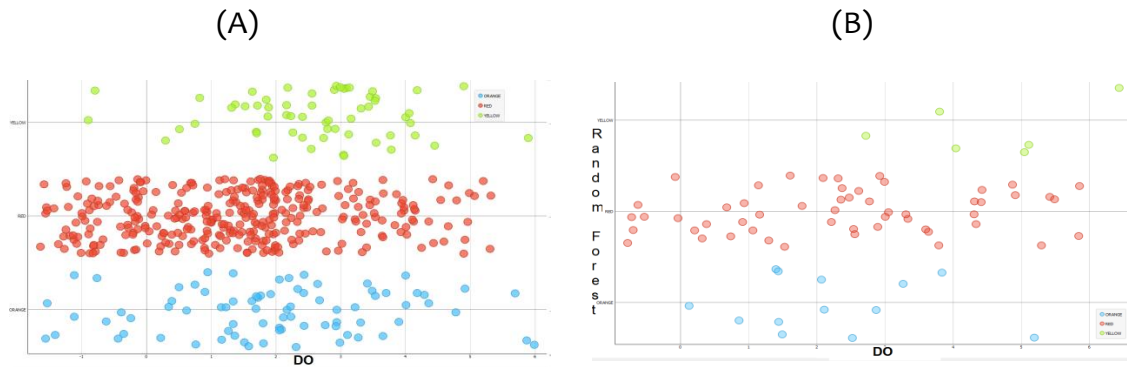

*Figure 6.* Scatter Plot for Dissolve Oxygen, (A) refers to the Training set and (B) refers to the prediction of the Random Forest

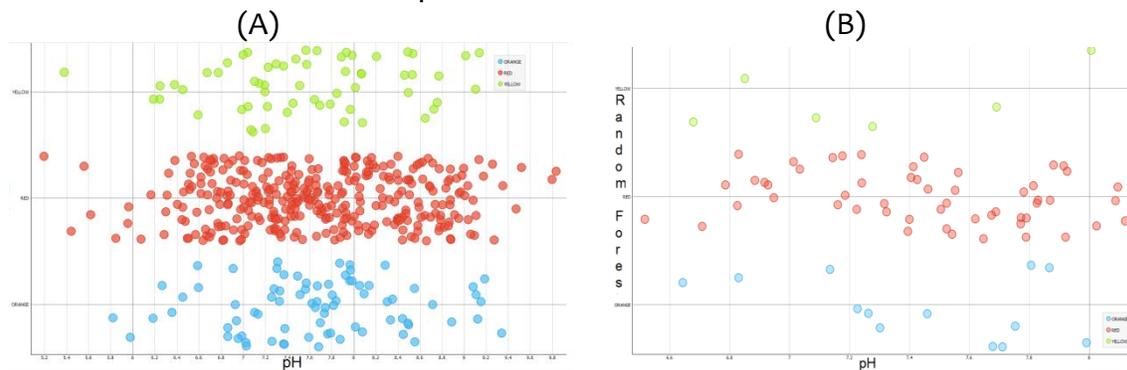

*Figure 7.* Scatter Plot for pH, (A) refers to the Training set and (B) refers to the prediction of the Random Forest

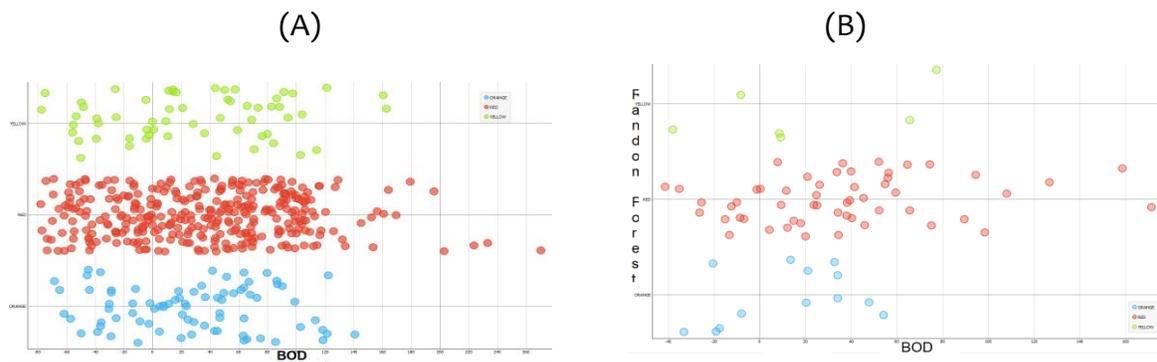

*Figure 8.* Scatter Plot for BOD, (A) refers to the Training set and (B) refers to the prediction of the Random Forest

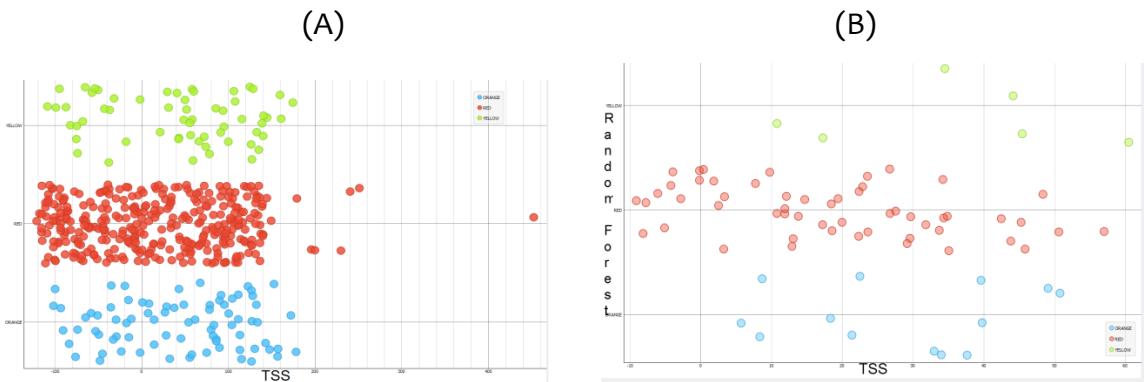

(A)                                    (B)

*Figure 9.* Scatter Plot for TSS, (A) refers to the Training set and (B) refers to the prediction of the Random Forest

After the execution of Random Forest Classifier, the model produces a correctly classified instance of 91.75% of accuracy and only a portion of incorrectly classified instances of 8.24% is shown in Table 4 below. The detailed accuracy by class shows the weighted average for Precision 91.7%, Recall scored 91.8% and F-Measure achieved 91.6% of accuracy.

Table 4. The correctness of Random Forest Classifier

| Classification | No of sample | Correctness in percentage |
|---|---|---|
| Correct | 434 | 91.7548 |
| Incorrect | 39 | 8.2452 |

Table 5. Detailed Accuracy by Class

| Class | TP Rate | FP Rate | Precision | Recall | F-Measure |
|---|---|---|---|---|---|
| Red | 0.967 | 0.188 | 0.926 | 0.967 | 0.946 |
| Yellow | 0.847 | 0.014 | 0.893 | 0.847 | 0.870 |
| Orange | 0.759 | 0.018 | 0.896 | 0.759 | 0.822 |
| Weighted Average | 0.918 | 0.138 | 0.917 | 0.918 | 0.916 |

Table 6. Confusion Matrix generated by Random Forest

| a | b | c | <--Classified as |
|---|---|---|---|
| 324 | 5 | 6 | a - Red |
| 8 | 50 | 1 | b - Yellow |
| 18 | 1 | 60 | c - Orange |

Through the use of confusion matrix generated as shown above in Table 6, it presents the performance of the classifier to recognize the tuples of different classes and to assess the interrater reliability the researchers derived a Kappa value of 0.8115 based on Table 2 the Kappa value and the level of the agreement are evidently Strong.

## CONCLUSIONS AND RECOMMENDATIONS

This study was able to present a working model in predicting the Marilao River pollution by utilizing the random forest classification and was able to train from data taken from DENR-EMB Region 3. The resulting accuracies of the predicted model scored 91.75% in terms of correctly classified instances and were able to generate 0.8115 Kappa values which indicate that the model used, produced a strong level of agreement. From 2013 to 2017 based on the data sampling provided by EMB, shows that the river is highly polluted. Several issues like the underestimation of the water parameter results were seen, issues which can be addressed by incorporating additional data to the training process and also by validating the resulting model on the different training set. The discretion on decisions regarding the prediction of the model is attributed to the DENR-EMB unit as they have more hands-on experience with regards to monitoring, restoring and protecting the environment under the existing Philippine laws.

The authors recommend that this study, visualizing a data-driven approach in providing pollution level estimation that is viable regardless of the water parameter of a particular river system and that the predictive modeling method, be implemented on various major river systems across the Philippines.

## ACKNOWLEDGEMENT

The authors acknowledge DENR-EMB Region 3 for generously providing the dataset of Marilao Meycauayan Obando River (MMOR).